\documentclass{article}

 \usepackage[preprint]{neurips_2026}

\usepackage[utf8]{inputenc} % allow utf-8 input
\usepackage[T1]{fontenc}    % use 8-bit T1 fonts
\usepackage{hyperref}       % hyperlinks
\usepackage{url}            % simple URL typesetting
\usepackage{booktabs}       % professional-quality tables
\usepackage{amsfonts}       % blackboard math symbols
\usepackage{nicefrac}       % compact symbols for 1/2, etc.
\usepackage{microtype}      % microtypography
\usepackage{xcolor}         % colors

\usepackage{adjustbox}
\usepackage{array}
\usepackage[table]{xcolor}
\usepackage{multirow}
\usepackage{amsmath}
\usepackage{graphicx}

\title{Budgeted LoRA: Distillation as Structured Compute Allocation for Efficient Inference}

\author{%
\makebox[\textwidth][c]{%
\begin{minipage}[t]{0.47\textwidth}
\centering
Mohammed Sabry \\
ADAPT Centre, Dublin City University, Ireland\\
\texttt{mohammed.sabry@adaptcentre.ie}
\end{minipage}
\hfill
\begin{minipage}[t]{0.47\textwidth}
\centering
Anya Belz \\
ADAPT Centre, Dublin City University, Ireland\\
\texttt{anya.belz@dcu.ie}
\end{minipage}
}%
}
\begin{document}

\maketitle

\begin{abstract}
We study distillation for large language models under explicit compute constraints, with the goal of producing student models that are not only cheaper to train, but structurally efficient at inference time. While prior approaches to parameter-efficient distillation, such as LoRA, reduce adaptation cost, they leave the dense backbone unchanged and therefore fail to deliver meaningful inference savings.
We propose \textbf{Budgeted LoRA}, a distillation framework that treats model compression as a \emph{structured compute allocation problem}. Instead of using a fixed student architecture, we introduce a global compute budget that sets the final target fraction of dense computation retained. Under this constraint, the model redistributes capacity across dense and low-rank pathways via (i) module-level dense retention coefficients, (ii) adaptive low-rank allocation, and (iii) post-training compression that selectively removes, approximates, or preserves dense components. This formulation yields a \emph{family of students} controlled by a single budget dial. Empirically, Budgeted LoRA matches standard LoRA perplexity at a moderate budget with a $1.74\times$ compressed-module speedup; at an aggressive budget it achieves a $4.05\times$ speedup with moderate perplexity degradation, and it preserves higher accuracy on function-style in-context learning probes. These results suggest that, under compute-constrained distillation, retaining behavior is less about matching perplexity or removing more parameters than it is about controlling how dense computation is transferred to low-rank pathways.
\end{abstract}

\section{Introduction}\label{sec:intro}
Large language models (LLMs) achieve strong performance by scaling model size and training compute, but this scaling comes at a significant cost for deployment. Distillation has emerged as a practical solution, transferring knowledge from large teacher models to smaller student models that are cheaper to serve. However, existing distillation pipelines \citep{azimi2024kdlorahybridapproachefficient,yang2025llmneoparameterefficientknowledge} primarily optimize for parameter count or training efficiency, and often fail to produce models that are \emph{structurally efficient} at inference time.

A central limitation arises from the widespread use of parameter-efficient fine-tuning methods such as LoRA \citep{hu2021loralowrankadaptationlarge}. While LoRA reduces the number of trainable parameters, it preserves the dense backbone of the model, leaving the dominant inference cost unchanged. As a result, LoRA-based distillation \citep{yang2025llmneoparameterefficientknowledge} yields students that are cheaper to train, but not necessarily cheaper to deploy.

In this work, we revisit distillation from a different perspective. Rather than treating the student architecture as fixed, we formulate distillation as a \emph{structured compute allocation problem}. The key idea is to treat dense and low-rank pathways as competing computational resources, and to adapt their respective contributions under a limited compute budget.

To this end, we introduce \textbf{Budgeted LoRA}, a distillation framework that imposes an explicit global constraint on dense computation. A budget parameter $F \in [0,1]$ specifies the endpoint of the dense-retention schedule. Under this constraint, the method is realized through (i) selective retention or suppression of dense components in individual linear projections (e.g., attention $q/k/v/o$ and MLP up/gate/down projections), (ii) adaptive low-rank allocation, and (iii) post-training compression, including dense removal and low-rank approximation.

This contrasts with prior approaches that progressively remove dense computation (e.g., PC-LoRA \citep{hwang2024pcloralowrankadaptationprogressive}), collapsing the model into a single low-rank regime. In contrast, our formulation yields a \emph{continuum of models} parameterized by the budget $F$ which enables a direct and controllable trade-off between efficiency and performance.

%We evaluate Budgeted LoRA in a distillation setting where a 12-layer model is compressed into a 6-layer student and trained using teachers of varying scale. Across all settings, Budgeted LoRA traces a clear Pareto frontier: it matches the perplexity of standard LoRA while significantly reducing training and inference cost. Beyond perplexity, we find that Budgeted LoRA substantially improves the preservation of in-context learning (ICL) behavior, suggesting that compute allocation during training plays a critical role in retaining higher-level capabilities.

We evaluate Budgeted LoRA in a setting where a fixed 6-layer student is initialized from a 12-layer 0.13B model, reflecting our core compression setting, and then distilled using teachers of varying scale to study how teacher strength affects the quality-efficiency trade-off. Across all settings, Budgeted LoRA traces a more favorable perplexity-cost Pareto frontier than standard LoRA: at similar perplexity, it reduces training and compressed-module inference cost, and by varying the budget 
$F$, it exposes lower-cost operating points with controlled perplexity degradation. Beyond perplexity, we find that Budgeted LoRA retains higher accuracy on function-style ICL probes adapted from the Function Vectors evaluation suite~\citep{todd2024functionvectorslargelanguage}, suggesting that budget-aware compute allocation during training can help retain this class of higher-level behavior.

Our contributions are as follows:
\begin{itemize}
\item We propose a new formulation of distillation as a structured compute allocation problem under a global budget constraint.
\item We introduce Budgeted LoRA, which combines controller-enforced dense retention with learned low-rank allocation, and produces deployment-efficient students via post-training compression.
\item We show that a single budget dial yields a controllable quality-efficiency trade-off, forming a Pareto frontier across training and inference cost.
\item We demonstrate that Budgeted LoRA retains substantially higher performance than standard LoRA-based distillation on function-style ICL probes.
\end{itemize}

%These results suggest that achieving efficient language models is not solely a matter of reducing parameters, but of learning how to structure and allocate computation during training.
%These results imply that, in this distillation setting, efficient distillation is not just parameter reduction, but controlled transfer of dense computation to low-rank pathways during training.

\section{Related Work} \label{sec:related_work}

\paragraph{Knowledge distillation for language models.}
Knowledge distillation (KD)~\citep{hinton2015distillingknowledgeneuralnetwork} transfers behavior from a large teacher to a smaller student, commonly by matching softened output distributions and, optionally, intermediate representations.
For Transformer language models, KD has been widely used to produce compact deployment-oriented variants (e.g., DistilBERT~\citep{sanh2020distilbertdistilledversionbert}, TinyBERT~\citep{jiao2020tinybertdistillingbertnatural}, MiniLM~\citep{wang2020minilmdeepselfattentiondistillation}).
More recently, KD has become a standard component of LLM post-training pipelines, where strong teachers are often too costly to serve directly and distilled students provide a practical quality–latency trade-off~\citep{mansourian2025comprehensivesurveyknowledgedistillation}.
However, most KD approaches assume a fixed student architecture and do not explicitly control how computation is allocated within the student.

\paragraph{LoRA and parameter-efficient distillation.}
Low-Rank Adaptation (LoRA)~\citep{hu2021loralowrankadaptationlarge} freezes pretrained weights and introduces trainable low-rank updates, enabling parameter-efficient adaptation while preserving the base network structure.
This makes LoRA a natural substrate for distillation when full student fine-tuning is expensive.
However, standard LoRA does not reduce inference compute: even when merged, the dense matrix multiplication remains unchanged, so LoRA-distilled students retain the full serving cost of the backbone.
Several recent works explore parameter-efficient distillation variants and hybrid KD objectives in LoRA settings~\citep{azimi2024kdlorahybridapproachefficient,yang2025llmneoparameterefficientknowledge}, but they primarily focus on reducing \emph{training cost}, leaving inference efficiency unaddressed.

\paragraph{Adaptive low-rank allocation.}
A growing body of work argues that a single global rank is suboptimal, and instead allocates rank across LoRA modules.
AdaLoRA \citep{zhang2023adaloraadaptivebudgetallocation} and related approaches dynamically redistribute rank based on estimated importance, typically under a parameter or rank budget.
These methods improve parameter efficiency, but their objective remains \emph{fine-tuning under parameter constraints}.
In particular, they do not couple rank allocation to changes in dense computation, and therefore do not directly optimize for inference efficiency.

\paragraph{Dense removal and progressive compression.}
Inference efficiency is often pursued via pruning and structured sparsification \citep{muralidharan2024compactlanguagemodelspruning,sanh2020movementpruningadaptivesparsity,liu2017learningefficientconvolutionalnetworks,louizos2018learningsparseneuralnetworks,wen2016learningstructuredsparsitydeep}, including approaches that learn gates to remove weights or structures with minimal quality loss.
PC-LoRA \citep{hwang2024pcloralowrankadaptationprogressive} is particularly relevant, as it progressively suppresses the pretrained dense pathway during training to enable adapter-only deployment.
However, PC-LoRA uses a single global decay schedule that drives a dense pathway toward zero uniformly, whereas our method formulates this process as an explicit compute-allocation problem with module-level dense retention under a global budget.

\paragraph{Positioning of our work.}
Our work lies at the intersection of these directions, but differs in one key way: we treat distillation as a \emph{structured compute allocation problem} and jointly control dense computation and low-rank capacity during training.
Concretely, our method differs from prior work in three ways. First, unlike AdaLoRA, we couple rank allocation with dense-path retention and removal under a shared deployment-oriented compute budget, rather than optimizing rank or parameter allocation alone. Second, unlike PC-LoRA, which progressively suppresses the dense pathway using a single global decay schedule, Budgeted LoRA enforces a global compute constraint while allowing individual projections—including attention $q/k/v/o$ and MLP up/gate/down projections—to selectively retain or remove dense capacity while jointly allocating low-rank capacity. This yields a family of students parameterized by the budget $F$, rather than a single compressed model. Third, we connect training-time allocation decisions to post-training compression, enabling multiple deployment modes: dropping dense components, approximating them via SVD, or retaining them when needed.

%\paragraph{Summary.}
%Across prior work, rank adaptation and dense removal are typically studied in isolation or under pure fine-tuning objectives. We unify them within a single \emph{budgeted distillation} framework, where a global compute budget governs how capacity is allocated during training and realized at inference. This results in students that are both efficient and capable, bridging the gap between parameter-efficient training and deployment efficiency.

\begin{figure*}[t]
\centering
\includegraphics[width=\textwidth]{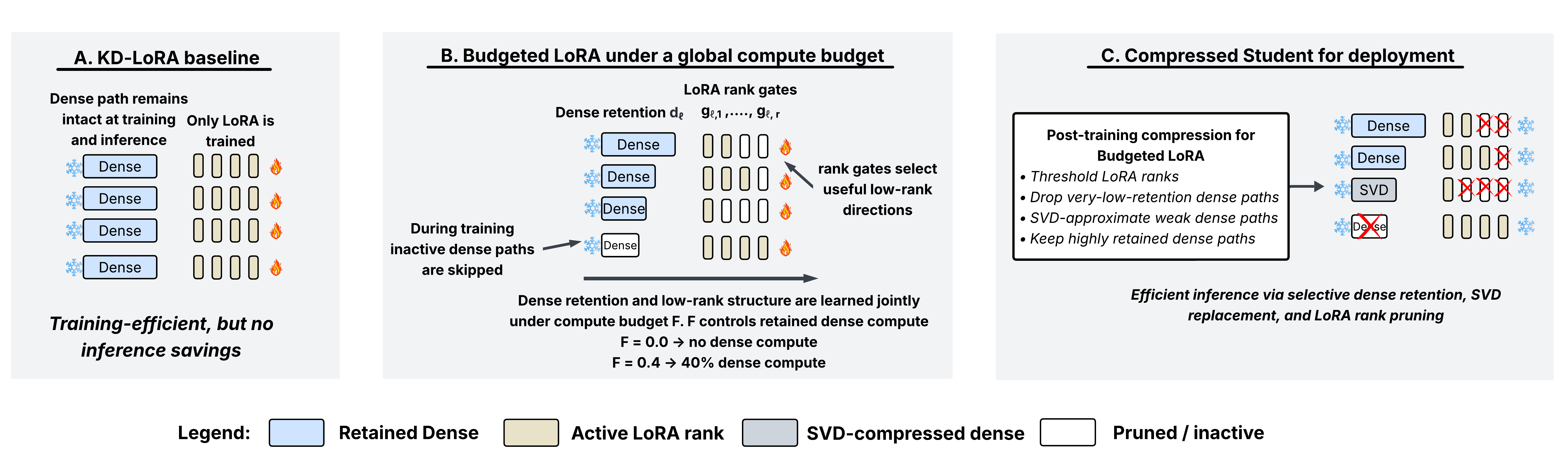}
\caption{
\textbf{Distillation as structured compute reallocation.}
\textbf{A:} Starting from KD-LoRA, only low-rank updates are trained, while the dense backbone remains unchanged at inference. 
\textbf{B:} Budgeted LoRA introduces a global compute budget realized through dense-retention coefficients on individual linear projections, together with learned LoRA rank gates during distillation. This enforces a compute allocation between dense and low-rank pathways, while the model adapts under the imposed budget.
\textbf{C:} After training, post-training compression consolidates the student into an efficient deployment form: LoRA ranks are thresholded, dense paths with very low retention are removed, weak dense paths are replaced by low-rank SVD approximations, and highly retained dense paths are kept.
}
\label{fig:budgeted_lora_pipeline}
\end{figure*}

\section{Method: Budgeted LoRA}
\label{sec:bud_distill_inference}

Figure~\ref{fig:budgeted_lora_pipeline} summarizes the three stages of our approach: the KD-LoRA starting point (A), budget-constrained distillation with dense-retention coefficients and rank gates (B), and post-training compression into a deployment-efficient student (C). 
This pipeline illustrates our central view: rather than treating the student structure as fixed, distillation under structural constraints can be formulated as a problem of allocating computation across dense and low-rank pathways.

%\paragraph{Method intuition.}
%As illustrated in Figure~\ref{fig:budgeted_lora_pipeline}, once LoRA is applied to a transformer layer, each adapted linear projection can be viewed as having two complementary computation pathways: the original dense pathway and an added low-rank pathway. Standard LoRA augments a dense path with a fixed low-rank update, but does not alter how computation is distributed between the two.

%In our method, a budget controller sets module-level dense \emph{retention} coefficients during training, while rank gates determine the effective capacity of the low-rank pathway. Together, these mechanisms determine the \emph{allocation} of computation between dense and low-rank components under a global budget constraint.

%By gradually reducing the retained dense budget over training, we enforce a controlled transition from dense-supported computation toward low-rank computation. In this view, the schedule links the compute constraint to a gradual reorganization of computation, allowing the student to adapt before dense capacity is removed.

%This perspective reframes distillation as structured training under compute constraints. Rather than allowing the model to freely allocate computation, we enforce a global budget over dense computation, and the model adapts its parameters within this constraint. As a result, the effective student structure is not predefined, but emerges as a function of the imposed budget, enabling a direct and continuous trade-off between efficiency and performance.

\paragraph{Method intuition.}
Once LoRA is applied to a Transformer projection, computation is split between the frozen dense pathway and the added low-rank pathway. Standard LoRA trains the low-rank update while leaving the dense pathway unchanged. Budgeted LoRA instead treats these pathways as competing computational resources: a controller sets module-level dense \emph{retention} coefficients, while rank gates determine the effective low-rank capacity. Gradually reducing the dense budget encourages a controlled transfer from dense-supported computation to low-rank pathways before post-training compression removes, approximates, or preserves dense components. The resulting student structure is therefore not fixed in advance, but emerges as a function of the imposed budget.

\paragraph{Method overview.}
We consider language-model distillation from a teacher to a fixed-depth student, with the goal of obtaining a deployment model that is both efficient to train and structurally efficient at inference time. 
Our method starts from a warm-started student constructed by selecting a subset of teacher layers, and replaces selected linear projections with budget-aware gated LoRA modules.

The model is trained with the standard distillation objective (KL divergence to the teacher logits plus cross-entropy with ground-truth labels; see Appendix~\ref{sec:ablation_kd}), while a budget controller dynamically adjusts dense retention to match a target compute trajectory over training. After training, we compress the resulting student into a deployment form that realizes actual inference savings.

We next describe the gated LoRA modules, budget controller, and post-training compression.

\subsection{Budget-aware Gated LoRA Modules} \label{sec:budget_aware_gated_lora_modules}

We replace selected linear projections in the student with budget-aware gated LoRA modules.
For an input $x \in \mathbb{R}^{d_{\mathrm{in}}}$, where $d_{\mathrm{in}}$ is the input dimension, and a base linear map $W \in \mathbb{R}^{d_{\mathrm{out}}\times d_{\mathrm{in}}}$, where $d_{\mathrm{out}}$ is the output dimension, each gated module computes
\begin{equation}
y
=
d \cdot Wx
+
\frac{\alpha}{r_{\max}} \, B \bigl((Ax) \odot g \bigr),
\label{eq:gatedlora}
\end{equation}
where $A \in \mathbb{R}^{r_{\max}\times d_{\mathrm{in}}}$ and $B \in \mathbb{R}^{d_{\mathrm{out}}\times r_{\max}}$ are the LoRA factors, $r_{\max}$ is the maximum configured LoRA rank, $g \in (0,1)^{r_{\max}}$ is a vector of learned per-rank gates, $d \in [0,1]$ is a dense-path retention coefficient controlled by the budget scheduler, and $\alpha$ is the usual LoRA scaling. In our Mistral-style experiments, the adapted projections are bias-free. The dense weight $W$ is frozen, while the LoRA factors and gate logits are trainable.

This parameterization of linear projections serves two purposes.
First, the rank gates allow different modules to use different effective ranks rather than sharing one global low-rank bottleneck.
Second, the dense retention coefficient controls how much of each dense projection remains active during training, making it possible to selectively and gradually reduce dense computation rather than committing to a hard structural change at initialization.
The result is a student whose trainable adaptation capacity and retained dense capacity can both vary across modules and over time.

\subsection{Budget Scheduling and Controller}
\label{sec:bud_scheduler_controller}
The dense budget is enforced by a controller operating over all gated modules.
For each module $m$, we define a dense cost proxy
\begin{equation}
c_m = d^{\mathrm{in}}_m \cdot d^{\mathrm{out}}_m,
\end{equation}
i.e.\ the MAC count of the dense linear projection (proportional to FLOPs up to a constant factor).
The total retained dense cost at training step $t$ is then
\begin{equation}
C(t) = \sum_{m=1}^{M} d_m(t)\, c_m,
\end{equation}
where $d_m(t)$ is the current retention coefficient value for the $m$th module.

We specify a target dense-budget fraction $b(t)$ using a cosine schedule.
Let $t_0$ and $t_1$ denote the start and end of the budgeting phase, expressed as fractions of total training steps, and let $F \in [0,1]$ be the final dense fraction.
Then
\begin{equation}
b(t)=
\begin{cases}
1, & t \le t_0,\\[2pt]
F + (1-F)\,\frac{1+\cos\!\bigl(\pi \frac{t-t_0}{t_1-t_0}\bigr)}{2}, & t_0 < t < t_1,\\[6pt]
F, & t \ge t_1.
\end{cases}
\end{equation}
The target dense cost is
\begin{equation}
C^\star(t)=b(t)\sum_{m=1}^{M} c_m.
\end{equation}

At each step, the controller enforces the scheduled dense-cost target by greedily lowering  module-level retention coefficients until the retained dense cost satisfies the budget. In our experiments, modules are assigned a removal score equal to their dense MAC cost, $s_m=c_m$, and are sorted in ascending score order, so lower-cost modules are reduced first. When modules have identical scores, ties follow the deterministic module registration order returned by \texttt{named\_modules()}.  The controller lowers retention coefficients in this order by the amount needed to satisfy the current budget target; these updates are then smoothed with an exponential moving average, and values below a small threshold are clamped to zero. Thus, the budget acts as an explicit control signal: early in training the student behaves close to a dense warm-started model, while later training progressively and selectively reallocates computation from dense paths to low-rank paths under a global compute constraint.

In summary, \emph{retention} refers to the training-time dense coefficient values, \emph{allocation} refers to the resulting distribution of compute across dense and low-rank pathways under the budget, and \emph{removal} is the hard zeroing of dense paths once retention falls below threshold.

\subsection{Post-Training Compression for Efficient Inference}
\label{sec:post_train_compress}
During training, a dense path is computed unless the controller has clamped its retention coefficient to zero, while the low-rank path is still evaluated at the full configured rank.
Thus, the raw training graph does not yet realize the full inference savings implied by the learned gates.
We therefore apply an explicit post-training compression pass before deployment. Here, \emph{compression} denotes the full post-training consolidation step, while \emph{removal} refers specifically to the case where a dense path is dropped entirely.

For each gated linear module, we first harden the rank gates by thresholding, and keep at least one active rank.
We then choose one of three deployment cases based on the final retention coefficient value $d$.
We use two retention thresholds to determine the deployment form: 
$\epsilon_{\mathrm{zero}}$ is the dense-removal threshold below which a dense path is dropped entirely, and $\epsilon_{\mathrm{lr}}$ is the low-rank approximation threshold below which the remaining dense contribution is approximated by SVD rather than kept as a full dense matrix.

\paragraph{Case 1: drop dense pathway entirely.}
If $d < \epsilon_{\mathrm{zero}}$, we remove a dense path and keep only the pruned low-rank component.
This yields a two-layer low-rank factorization implementing the LoRA path alone.

\paragraph{Case 2: approximate residual dense contribution by low-rank SVD.}
If $\epsilon_{\mathrm{zero}} \le d < \epsilon_{\mathrm{lr}}$, we treat the remaining dense contribution $dW$ as small enough to approximate with a low-rank SVD.
We compute
\begin{equation}
dW \approx U_k V_k,
\end{equation}
with rank $k$ chosen adaptively up to a configured maximum.
Specifically, we set
\begin{equation}
k = \max\left(1,\left\lfloor r_{\max}^{\mathrm{dense}} \cdot 
\frac{d}{\epsilon_{\mathrm{lr}}}\right\rceil\right),
\end{equation}
where $r_{\max}^{\mathrm{dense}}$ is the maximum SVD rank allowed for approximating residual dense components; we set $r_{\max}^{\mathrm{dense}}=128$ in all experiments.
The rank is clipped to at most $\min(d_{\mathrm{out}}, d_{\mathrm{in}})$.
Thus, dense components with larger retained coefficients receive higher SVD rank, while components closer to the removal threshold receive lower rank.
We then concatenate this dense approximation with the hardened LoRA factors, yielding a single fused low-rank module.
This preserves some residual dense information while avoiding a full dense matrix multiply.

\paragraph{Case 3: keep dense pathway and merge.}
If $d \ge \epsilon_{\mathrm{lr}}$, we retain a dense path and merge it with the hardened LoRA update into a standard dense linear layer:
\begin{equation}
W_{\mathrm{eff}} = dW + \Delta W_{\mathrm{LoRA}}.
\end{equation}
This case is used when the module still appears to require dense capacity for accuracy.

These three cases produce a deployment model that is materially different from the training graph.
In particular, compression can realize both rank pruning and dense removal, and therefore translate training-time budgeting signals into actual inference-time savings.
Because the resulting modules are standard dense or sequential low-rank operators, the compressed student can be evaluated and served without the original gating machinery.

%Furthermore, we observe a dynamic coordination between dense retention and rank gating across budgets (Table~\ref{tab:budlora-compression}). When the budget is low, post-training compression is dominated by Case 1 (dense drop), so the LoRA pathway carries most of the computation and the effective rank remains near its maximum. As the budget increases, fewer modules are fully dropped and more fall into Case 2 or Case 3, reducing pressure on the LoRA pathway and allowing additional rank directions to be removed.

%Notably, this coordination emerges under a fixed set of hyperparameters and without auxiliary losses: varying only the global budget $F \in \{0.0, 0.4, 0.8\}$ is sufficient to induce distinct compression regimes. This behavior is analyzed in Appendix~\ref{sec:dynamic-compression}.

The interaction between dense retention, rank gating, and post-training compression is analyzed in Appendix~\ref{sec:dynamic-compression}.

\section{Experimental Setup}
\label{sec:experiments}

\begin{table}[t]
\centering
\small
\setlength{\tabcolsep}{3.5pt}
\renewcommand{\arraystretch}{0.95}
\begin{tabular}{@{}lccccc@{}}
\toprule
\textbf{Teacher} & \textbf{Layers} & $\boldsymbol{d_{\mathrm{model}}}$ & $\boldsymbol{d_{\mathrm{ff}}}$ & \textbf{Heads/KV} & \textbf{PPL $\downarrow$} \\
\midrule
0.13B & 12 & 768  & 3072 & 12/3 & 21.8 \\
0.5B  & 30 & 1024 & 4096 & 16/4 & 16.0 \\
1B    & 30 & 1536 & 6144 & 24/6 & 14.1 \\
\bottomrule
\end{tabular}
\caption{Teacher checkpoints used for distillation. All use a Mistral-style decoder-only architecture with head dimension 64 and are pretrained on \textsc{The Pile} \citep{gao2020pile800gbdatasetdiverse} under the same compute-optimal setup; PPL denotes validation perplexity.}
\label{tab:teacher-models}
\end{table}

\subsection{Models, Data, and Evaluation} \label{sec:exp_setup} 

\paragraph{Models.}
All teacher and student models use a Mistral-style decoder-only Transformer architecture \citep{jiang2023mistral7b}, with pre-norm blocks, rotary position embeddings (RoPE), SwiGLU feed-forward layers, and grouped-query attention (GQA). 
The checkpoints are taken from the Bi-Induct model family \citep{sabry2026inductionsignaturesenoughmatchedcompute}. 
We evaluate distillation across three teacher scales: 0.13B, 0.5B, and 1B. 
All teacher models were pretrained on \textsc{The Pile} \citep{gao2020pile800gbdatasetdiverse} under a Chinchilla-style compute-optimal setup \citep{hoffmann2022trainingcomputeoptimallargelanguage}, yielding validation perplexities of 21.8, 16.0, and 14.1, respectively (Table~\ref{tab:teacher-models}). 
This provides a controlled ladder of teacher quality. The deployment student is always a fixed 6-layer model initialized from the 0.13B checkpoint; for the 0.5B and 1B settings, only the teacher scale changes. This isolates distillation effects from differences in student initialization.

\paragraph{Data.}
Teacher pretraining and student distillation use the same deduplicated corpus (The Pile \citep{gao2020pile800gbdatasetdiverse}), with identical preprocessing and partitioning. 
All data are tokenized in the same way and truncated to sequences of length 1024. 
This shared pipeline ensures that differences between teachers, truncated students, and distilled students are not confounded by data distribution or data handling.

We define a fixed held-out evaluation slice using a stable MD5-based hash, corresponding to 0.2\% of the corpus (roughly 0.4B tokens). 
All evaluation subsets are drawn from this slice to ensure consistency across runs.

\paragraph{Training.}
All student models are trained for 2.55B tokens with sequence length 1024.
This matches the pretraining token budget of the 0.13B model, enabling a direct comparison between:
(i) a supervised model trained from scratch on the same training corpus and for the same number of tokens, and
(ii) its distilled or truncated counterparts.
All final distillation runs are performed on two NVIDIA A100 GPUs using bf16 precision, with per-device batch size of $2^{14}$ tokens.
We use AdamW with a cosine learning-rate schedule, base learning rate $3\times 10^{-4}$, zero weight decay, and 3\% warmup capped at 2000 steps.
Gradients are clipped to a maximum norm of 1.0.

\paragraph{Evaluation.}
Across all models, language modeling quality is measured using perplexity on the same 10M-token validation set drawn from the held-out \textsc{The Pile} evaluation slice.

We additionally evaluate ICL using a suite of 19 function-style probes adapted from the Function Vectors evaluation suite \cite{todd2024functionvectorslargelanguage} and used in prior analyses of ICL circuits \citep{sabry2026inductionsignaturesenoughmatchedcompute}. 
These probes require the model to infer a simple input-output mapping from demonstrations, such as selecting an item from a sequence, changing capitalization, or predicting the next item. 
They are chosen because they depend strongly on using the provided context, reducing confounds from alternative pathways such as memorized facts or parametric recall. 
We follow a 10-shot prompting protocol with three random demonstration seeds; full task and prompting details are provided in Appendix~\ref{app:todd-tasks}. We report exact-match accuracy on these probes and interpret the results as relative retention of ICL-probe behavior rather than as evidence of high absolute task competence.

\subsection{Compared Methods}
\label{sec:compared_methods}

We compare our proposed method against two baseline families (full-parameter KD and KD-LoRA), capturing different efficiency trade-offs. Unless otherwise stated, all hyperparameters and architectural choices are fixed according to the ablation study in Section~\ref{app:ablations}.

\paragraph{KD Full.}
This baseline trains a 6-layer student with full-parameter distillation. 
The student is warm-started using the \texttt{mixed} layer-selection scheme, which we found to consistently outperform other initialization strategies (Section~\ref{sec:ablation_kd}). 
Training uses a weighted combination of logit KD and next-token cross-entropy. 
Following the ablation results, we use temperature $\tau_{\mathrm{KD}}=3.0$ and KD mixing coefficient $\lambda_{\mathrm{KD}}=0.8$.

\paragraph{KD LoRA.}
This baseline keeps the same 6-layer warm-started student construction, but replaces full tuning with LoRA-based parameter-efficient distillation. 
We apply LoRA to all linear layers 
(\texttt{q\_proj}, \texttt{k\_proj}, \texttt{v\_proj}, \texttt{o\_proj}, \texttt{gate\_proj}, \texttt{up\_proj}, \texttt{down\_proj}), 
resulting in 42 adapted modules. 
Based on ablations (Section~\ref{sec:ablation_lora}), we fix the rank to $r=128$, with scaling $\alpha=256$ and dropout $0$. 
As with KD Full, we use $\tau_{\mathrm{KD}}=3.0$ and $\lambda_{\mathrm{KD}}=0.8$, keeping these fixed across teacher scales for controlled comparison. 
This baseline reduces adaptation cost, but standard LoRA does not by itself reduce dense inference cost.

\paragraph{Budgeted LoRA (ours).}
Our method (described in detail in Section~\ref{sec:bud_distill_inference}) builds on KD LoRA by introducing adaptive rank allocation and dense-path removal under an explicit budget schedule.
Unless otherwise stated, we use the same hyperparameters as KD LoRA, including $\tau_{\mathrm{KD}}=3.0$, $\lambda_{\mathrm{KD}}=0.8$, and rank $r=128$, as these transfer robustly across methods (Section~\ref{sec:ablation_lora}). 
Budget scheduling and compression thresholds are set based on the ablation study (Section~\ref{sec:ablation_budgeted_lora}), with short decay schedules and thresholds selected to balance performance and efficiency. 
We report its quality, ICL retention, training cost, and compressed inference cost in the final evaluation.

\section{Results} \label{sec:all_results}
We evaluate Budgeted LoRA along two axes: quality-efficiency trade-offs measured by perplexity, training cost, and inference cost (Section~\ref{sec:main-quality-efficiency}), and In-Context Learning retention (Section~\ref{sec:icl_results}).

\subsection{Quality–Efficiency Trade-offs}
\label{sec:main-quality-efficiency}
Table~\ref{tab:main-quality-efficiency} summarizes validation perplexity together with compact student-side efficiency metrics across the three teacher scales. 
The adapted-module training-cost proxy is defined in Appendix~\ref{app:compute}; it measures scheduled compute allocation rather than exact hardware FLOPs or wall-clock training time.

\paragraph{Baselines.}
Among the baselines, full-parameter KD consistently achieves the best perplexity and improves as teacher scale increases. Notably, it also outperforms the 0.13B supervised reference model from which the students are initialized (validation perplexity 21.8; Table~\ref{tab:teacher-models}).

By contrast, KD LoRA exhibits a persistent gap relative to full KD across all teacher scales and does not benefit from larger teachers in the same way. One plausible explanation is a distillation capacity gap \citep{busbridge2025distillationscalinglaws}: as teacher scale increases, the supervision signal becomes richer, but LoRA-based students can only adapt through a limited low-rank subspace around a frozen backbone. Full-parameter KD can exploit stronger teachers because the entire student can reorganize, whereas LoRA-based methods may be unable to absorb the additional information as effectively.

\paragraph{Budgeted LoRA trade-off.}
Budgeted LoRA introduces a controllable trade-off through the dense-compute budget $F$. 
At $F{=}0.4$, it closely matches KD LoRA in perplexity across all scales (e.g., 24.34 vs.\ 24.32 at 0.13B), while substantially reducing both training and compressed-module inference cost (0.59$\times$ training compute and 1.74$\times$ compressed-module inference speedup). 
At the more aggressive setting $F{=}0.0$, Budgeted LoRA yields the largest efficiency gains (0.38$\times$ training compute and 4.05$\times$ compressed-module inference speedup), at the cost of a moderate degradation in perplexity.

\paragraph{Pareto frontier.}
Overall, Budgeted LoRA traces a clear Pareto frontier: decreasing $F$ yields substantial improvements in efficiency while incurring a controlled loss in perplexity. 
This demonstrates that a single global budget parameter is sufficient to smoothly navigate the quality-efficiency trade-off in distillation.

%\paragraph{Bridge to behavior.}
%While these results characterize the efficiency-quality trade-off, they do not reveal whether distilled models preserve functional behavior. We turn to this question in the next section.

The extreme setting $F{=}0.0$ is the low-rank-only endpoint of our compute-allocation spectrum, where retained dense computation is driven to zero after compression. This endpoint provides the closest comparison to progressive dense-removal approaches such as PC-LoRA~\citep{hwang2024pcloralowrankadaptationprogressive} within our distillation setup. In our formulation, however, full dense removal is not a separate method, but one operating point of a budgeted family: varying $F$ yields a continuum between dense and low-rank computation rather than committing to full dense removal a priori.

\begin{table*}[t]
\centering
\small
\begin{tabular}{lccc|ccc}
\toprule
& \multicolumn{3}{c|}{\textbf{Quality (PPL $\downarrow$)}} & \multicolumn{3}{c}{\textbf{Efficiency (student-side)}} \\
\cmidrule(lr){2-4} \cmidrule(lr){5-7}
\textbf{Method} & 0.13B & 0.5B & 1B & Train Compute $\downarrow$ & Infer Speedup $\uparrow$ & Param Red. $\uparrow$ \\
\midrule

\multicolumn{7}{c}{\textit{Baselines}} \\
\midrule
KD Full & \textbf{20.88} & \textbf{18.26} & \textbf{17.86} & $1.00\times$ & $1.00\times$ & 0\% \\
KD LoRA & 24.32 & 26.06 & 26.50 & $0.91\times$ & $1.00\times$ & 0\% \\

\midrule
\multicolumn{7}{c}{\textit{Budgeted LoRA (ours)}} \\
\midrule
$F = 0.4$ & 24.34 & 26.19 & 26.58 & $0.59\times$ & $1.74\times$ & 53.9\% \\
$F = 0.0$ & 25.13 & 28.45 & 28.86 & \textbf{0.38$\times$} & \textbf{4.05$\times$} & \textbf{80.2\%} \\

\bottomrule
\end{tabular}
\caption{
Quality-efficiency tradeoffs across distillation methods. Lower perplexity is better. 
Budgeted LoRA exposes a controllable tradeoff via the dense budget $F$: lower $F$ improves student-side training compute and compressed-module inference speedup at the cost of higher perplexity. Training compute is an adapted-module allocation proxy normalized to KD Full and excludes amortized teacher cost. Parameter reduction is relative to the pre-compression footprint of the gated modules.
}
\label{tab:main-quality-efficiency}
\end{table*}

\begin{figure*}[t]
\centering
\includegraphics[width=0.93\textwidth]{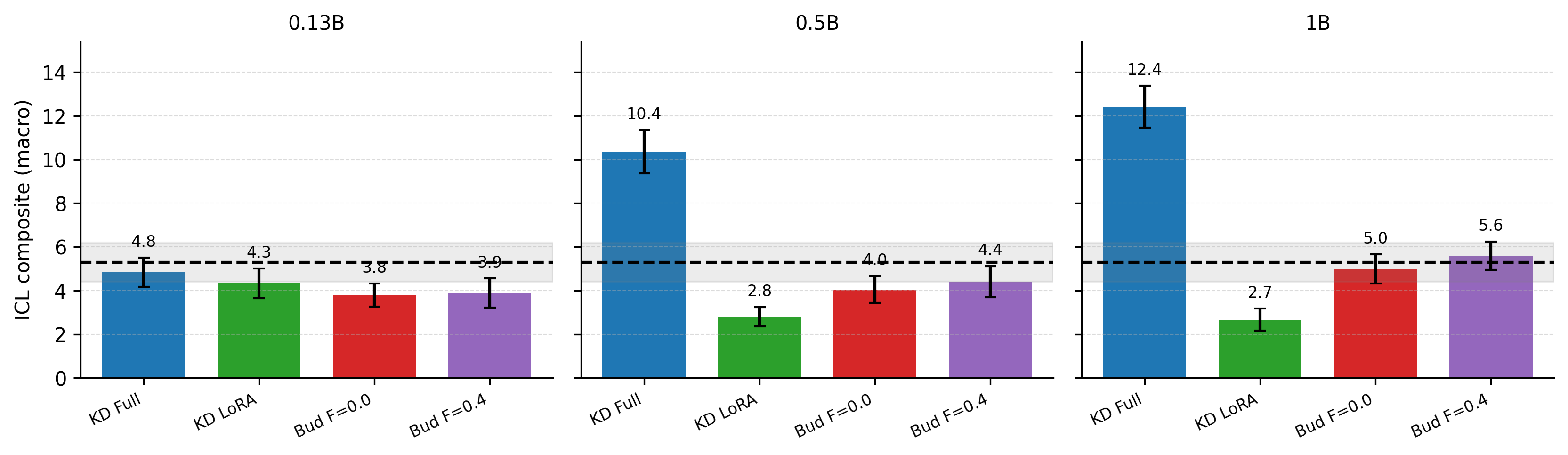}
\caption{
Macro-averaged ICL composite accuracy across 19 function-style probes adapted from the Function Vectors evaluation suite \citep{todd2024functionvectorslargelanguage} for students distilled from teachers at three scales (0.13B, 0.5B, and 1B). Bars show the mean across tasks and three random demonstration seeds, with error bars indicating variability across demonstration seeds. The dashed horizontal line marks the score of the full 0.13B model used as the student initialization source. Budgeted LoRA (\(F\in\{0.0,0.4\}\)) retains higher probe accuracy than vanilla KD-LoRA at larger teacher scales, while remaining below full-parameter KD.
}
\label{fig:icl-composite}
\end{figure*}

\subsection{In-Context Learning Retention}
\label{sec:icl_results}
Figure~\ref{fig:icl-composite} shows the macro-averaged ICL composite score across probes (mean accuracy). The dashed horizontal line marks the performance of the full 0.13B initialization model, and the shaded region indicates variability across runs.
A detailed per-task breakdown is provided in Table~\ref{tab:main_icl_performance}.

Notably, methods with similar perplexity (Table~\ref{tab:main-quality-efficiency}) can differ substantially on ICL probes; given the low absolute accuracies of these small models, we interpret the results as relative probe retention.

\paragraph{Main results.}
Several trends emerge. 
First, KD Full benefits strongly from increasing teacher scale, with ICL performance improving substantially from 0.13B to 1B. 
Because KD Full updates all student parameters, it can better absorb the stronger teacher signal, but it is also the least efficient distillation setting in our comparison, with full student-side training cost and no compressed-module inference speedup (Table~\ref{tab:main-quality-efficiency}).
In contrast, KD LoRA degrades on these probes as teacher scale increases, indicating weaker relative retention under standard parameter-efficient KD.

Budgeted LoRA substantially mitigates this failure mode. 
Across both 0.5B and 1B teachers, it consistently outperforms KD LoRA, recovering part of the probe accuracy lost under standard LoRA. For example, at 1B, KD LoRA achieves 2.7, while Budgeted LoRA reaches 5.0 ($F{=}0.0$) and 5.6 ($F{=}0.4$).

Interestingly, Budgeted LoRA exhibits a different scaling behavior across metrics. 
While its perplexity follows the same trend as KD LoRA and does not improve with larger teachers (Table~\ref{tab:main-quality-efficiency}), its ICL performance improves substantially with teacher scale, mirroring the behavior of full-parameter KD. 
This decoupling suggests that preserving higher-level behavior does not require fully matching the teacher in perplexity, but instead depends on maintaining the underlying computational structure.

\paragraph{Effect of the budget.}
The comparison between budget settings reveals a consistent pattern: $F{=}0.4$ outperforms $F{=}0.0$ at larger teacher scales, suggesting that retaining a nonzero fraction of dense computation is necessary to preserve ICL-relevant structure. 
Thus, while lower budgets maximize efficiency, higher budgets better maintain behavioral capabilities.

\paragraph{Interpretation.}
Taken together with Table~\ref{tab:main-quality-efficiency}, these results show that the quality-efficiency tradeoff extends beyond perplexity. 
Budgeted LoRA not only improves efficiency, but also retains substantially higher probe accuracy than KD LoRA, especially when distilling from larger teachers.

This partially aligns with the distillation capacity gap observed in Section~\ref{sec:main-quality-efficiency}: constraining adaptation to a low-rank subspace limits the ability of the student to absorb the full distributional signal of stronger teachers, as reflected in perplexity. 
However, the improvement on these probes suggests that Budgeted LoRA may retain behavior-relevant structure that is not captured by perplexity alone.

%More broadly, these results provide strong evidence that preserving behavior is not solely a function of parameter count, but depends on how computation is structured during training. ICL behavior depends on structured internal circuits, such as induction-like patterns \citep{olsson2022incontextlearninginductionheads}, and crucially on whether those circuits are load-bearing \citep{sabry2026inductionsignaturesenoughmatchedcompute}. These mechanisms are sensitive to representational capacity and layer composition, and are therefore disrupted by abrupt or uniform compression, as in standard LoRA.

%In contrast, enforcing a compute budget during training encourages the model to preserve the critical dense pathways required for this behavior, resulting in stronger retention of ICL capability.

More broadly, these results suggest that preserving behavior is not solely a function of parameter count, but also of how computation is structured during training. ICL has been linked to structured internal mechanisms, including induction-like patterns~\citep{olsson2022incontextlearninginductionheads} and load-bearing circuit structure~\citep{sabry2026inductionsignaturesenoughmatchedcompute}. From this perspective, abrupt or uniform compression can disrupt behavior-relevant structure, while budgeted dense retention provides a smoother path for preserving it.

\section{Conclusion}
\label{sec:conclusion}

We have studied distillation under explicit compute constraints, arguing that efficiency should be viewed not only as a parameter-count problem, but as a problem of compute allocation. While standard LoRA reduces adaptation cost, it leaves the dense backbone largely unchanged and therefore fails to deliver meaningful inference savings.

We proposed \textit{Budgeted LoRA}, which enforces a global budget over dense computation during distillation and converts the resulting structure into an efficient deployment model. Across teacher scales, Budgeted LoRA traces a clear quality-efficiency Pareto frontier, substantially reducing training and compressed-module inference cost with controlled degradation in perplexity.

Beyond perplexity, we showed that compute allocation also affects capability retention. Although Budgeted LoRA follows LoRA-like perplexity trends across teacher scale, it behaves more like full-parameter KD on our function-style ICL probes, retaining substantially higher probe accuracy than standard LoRA. This suggests that higher-level behavior depends not only on matching the teacher’s output distribution, but also on the organization of the underlying computational structure.

Our analysis further shows that successful compression depends on how compute is controlled during training: gradual budgeted reduction enables a smoother transfer of capacity to low-rank pathways than abrupt dense removal. Overall, Budgeted LoRA reframes distillation as structured training under compute constraints, with a single global budget yielding cheaper training, compressed-module inference speedups, and stronger ICL-probe retention than standard LoRA-based distillation. %We discuss limitations, including hyperparameter transfer, module-level rather than end-to-end latency evaluation, and scaling to larger architectures, in Appendix~\ref{app:limitations}.

%Our analysis further showed that successful compression depends on how compute is controlled during training: abrupt dense removal leads to poorer solutions, whereas gradual budgeted reduction enables a smoother transfer of capacity to low-rank pathways.

%Overall, Budgeted LoRA reframes distillation as structured training under compute constraints: a single global budget parameter yields cheaper training, compressed-module inference speedups, and stronger ICL retention than standard LoRA-based distillation. We hope this perspective will be useful for future work on modularity, compression, and scalable adaptation in large language models.

\bibliography{neurips_2026}
\bibliographystyle{abbrv}

\appendix
\section{Limitations}
\label{app:limitations}
We do not yet establish transfer laws for Budgeted LoRA hyperparameters. Our ablations identify effective schedules, retention thresholds, and compression thresholds in the regimes studied here, but their scaling across budget levels, student sizes, teacher scales, and architectures remains open. We also use a fixed KD temperature and mixing coefficient across teacher scales to keep comparisons controlled; however, larger teachers may benefit from different distillation temperatures, and tuning these per scale may improve both KD-LoRA and Budgeted LoRA. In addition, our inference-speed measurements are computed over replaced/compressed modules rather than full end-to-end generation latency, and we have not yet evaluated very large-scale models or alternative architectures.

\section{Broader Impacts}
\label{app:broader_impacts}
Budgeted LoRA may reduce the cost of adapting and deploying language models, potentially lowering energy use and making efficient models more accessible. At the same time, cheaper adaptation and inference may also lower the barrier to misuse, including spam, misinformation, or other harmful deployments of language models. This work is intended as research on model compression and does not introduce application-specific safeguards.

\section{Ablation Studies}
\label{app:ablations}

We conduct a structured ablation study to isolate the key design choices underlying our approach. 
The ablations are organized progressively: we first establish strong distillation baselines, then introduce LoRA, and finally analyze the additional components specific to Budgeted LoRA.

For most ablation experiments, we use \(1/40\) of the full distillation token budget, since broad hyperparameter exploration at the full 2.55B-token scale would be prohibitively expensive. 
The main exception is budget-scheduler tuning for Budgeted LoRA, where we use the full token budget. Scheduler behavior is strongly token-dependent, and we did not establish a reliable transfer law from short runs to full-budget training; accordingly, tuning it directly at the target budget is more faithful than using a reduced-budget proxy.

All ablations use the same optimization setup as the main experiments, including batch size, learning rate, optimizer, and warmup schedule, as well as the same held-out validation set as the final models, ensuring that model-selection decisions are guided by the same evaluation signal used in the main experiments.

\subsection{Distillation Hyperparameters and Student Initialization}
\label{sec:ablation_kd}

\paragraph{Distillation Objectives}

Let $x=(x_1,\dots,x_n)$ be an input sequence.
The teacher produces logits $z^T_t \in \mathbb{R}^{V}$ and the student produces logits $z^S_t \in \mathbb{R}^{V}$ for each next-token prediction step $t$.
Our base distillation objective combines logit KD with the standard language-modeling objective:
\begin{align}
\mathcal{L}_{\mathrm{KD}}
&=
\frac{\tau_{\mathrm{KD}}^2}{|\Omega|}
\sum_{t \in \Omega}
\mathrm{KL}\!\left(
\mathrm{softmax}\!\left(\frac{z^T_t}{\tau_{\mathrm{KD}}}\right)
\,\middle\|\,
\mathrm{softmax}\!\left(\frac{z^S_t}{\tau_{\mathrm{KD}}}\right)
\right), \\
\mathcal{L}_{\mathrm{CE}}
&=
-\frac{1}{|\Omega|}
\sum_{t \in \Omega}
\log p_S(x_{t+1}\mid x_{\le t}).
\end{align}
Here, $\Omega$ is the set of non-padding positions and $\tau_{\mathrm{KD}}$ is the distillation temperature.
We include the standard $\tau_{\mathrm{KD}}^2$ factor to preserve the relative gradient scale of the KL term across temperatures.

The main training loss is:
\begin{equation}
\mathcal{L}_{\mathrm{base}}
=
\lambda_{\mathrm{KD}} \mathcal{L}_{\mathrm{KD}}
+
(1-\lambda_{\mathrm{KD}})\mathcal{L}_{\mathrm{CE}}.
\end{equation}

Using this objective, we first study the effect of the core distillation choices: the temperature $\tau_{\mathrm{KD}}$, the distillation weight $\lambda_{\mathrm{KD}}$, and the student initialization strategy. 
All experiments in this stage use a 0.13B teacher and a 6-layer student constructed via truncation (taking the first six layers).

We explore the following search space:
\begin{itemize}
    \item Initialization: \texttt{warmstart}, \texttt{scratch}
    \item Temperature: $T \in \{1.0, 2.0, 3.0, 4.0\}$
    \item Distillation weight: $\lambda \in \{0.3, 0.5, 0.7, 0.9\}$
\end{itemize}

We find that warmstarting the student from a truncated checkpoint consistently outperforms training from scratch, and the best distillation configuration is $T{=}2.0$, $\lambda{=}0.5$ (Table~\ref{tab:kd_temps_lams_warmstart_scratch}), which we use as a starting point for further exploration.

\begin{table}[t]
\centering
\small
\setlength{\tabcolsep}{6pt}
\renewcommand{\arraystretch}{1.12}
\begin{tabular}{cccc}
\toprule
Temperature $T$ & $\lambda_{\mathrm{KD}}$ & Warmstart PPL $\downarrow$ & Scratch PPL $\downarrow$ \\
\midrule
1.0 & 0.3 & 34.96 & 53.01 \\
2.0 & 0.5 & \textbf{34.08} & 52.61 \\
3.0 & 0.7 & 34.50 & 62.40  \\
4.0 & 0.9 & 37.39 & 85.08 \\
\bottomrule
\end{tabular}
\caption{The Pile Validation perplexity (PPL) for different distillation hyperparameter settings, comparing warmstart and scratch initialization. Lower is better.}
\label{tab:kd_temps_lams_warmstart_scratch}
\end{table}

\paragraph{Student construction.}
We next refine the warmstart strategy by varying how layers are selected from the teacher. 
We compare four variants: \textit{first}, \textit{middle}, \textit{last}, and \textit{mixed}. 
The first, middle, and last variants select contiguous blocks from the corresponding depth regions of the teacher. 
The mixed strategy instead anchors the student at the first and last teacher layers, then fills the remaining slots with approximately evenly spaced interior layers across depth. 
This provides coverage of early, middle, and late representations while preserving both low-level and high-level processing stages. 
Empirically, the mixed strategy consistently yields the best performance, suggesting that a depth-diverse initialization is more effective than copying a single contiguous block (Table~\ref{tab:kd_student_build_modes}).

\begin{table}[t]
\centering
\small
\setlength{\tabcolsep}{6pt}
\renewcommand{\arraystretch}{1.12}
\begin{tabular}{lc}
\toprule
Student construction mode & PPL $\downarrow$\\
\midrule
truncated & 34.08 \\
\textbf{mixed} & \textbf{28.13}  \\
middle & 29.58  \\
last & 32.15 \\
\bottomrule
\end{tabular}
\caption{The Pile Validation perplexity (PPL) for different student build modes under warmstart KD Full with fixed distillation hyperparameters ($T{=}2.0$, $\lambda_{\mathrm{KD}}{=}0.5$). Lower is better.}
\label{tab:kd_student_build_modes}
\end{table}

We then perform a second sweep over $(T,\lambda)$ for the top student constructions (\textit{mixed} and \textit{middle}), using:
\[
T \in \{1.0, 2.0, 3.0\}, \quad \lambda \in \{0.3, 0.5, 0.8\}.
\]
The best configuration is $T{=}3.0$, $\lambda{=}0.8$ with the mixed initialization (Table ~\ref{tab:kd_mixed_middle_temps_lams}). 
This configuration is fixed for all subsequent experiments.

\begin{table}[t]
\centering
\small
\setlength{\tabcolsep}{6pt}
\renewcommand{\arraystretch}{1.12}
\begin{tabular}{cccc}
\toprule
Temperature $T$ & $\lambda_{\mathrm{KD}}$ & Mixed PPL $\downarrow$ & Middle PPL $\downarrow$ \\
\midrule
1.0 & 0.3 & 29.01 & 30.55 \\
2.0 & 0.5 & 28.11 & 29.58 \\
3.0 & 0.8 & \textbf{27.84} & 29.92 \\
\bottomrule
\end{tabular}
\caption{The Pile Validation perplexity (PPL) for warmstart KD Full under two student build modes, mixed and middle, across different distillation hyperparameter settings. Lower is better.}
\label{tab:kd_mixed_middle_temps_lams}
\end{table}

\subsection{LoRA Design Choices}
\label{sec:ablation_lora}

We next introduce LoRA and study its additional design choices. 
We attach LoRA modules to all linear layers in the student (7 per layer across 6 layers, totaling 42 modules: \texttt{q\_proj}, \texttt{k\_proj}, \texttt{v\_proj}, \texttt{o\_proj}, \texttt{down\_proj}, \texttt{gate\_proj}, \texttt{up\_proj}).

We sweep the LoRA rank:
\[
r \in \{16, 32, 64, 128\}.
\]

We find the best-performing configuration uses rank $r{=}128$ (Table ~\ref{tab:kd_lora_rank_sweep}), which we fix for all LoRA-based experiments.

\begin{table}[t]
\centering
\small
\setlength{\tabcolsep}{6pt}
\renewcommand{\arraystretch}{1.12}
\begin{tabular}{ccc}
\toprule
LoRA rank $r$ & LoRA $\alpha$ & PPL $\downarrow$ \\
\midrule
16  & 32  & 39.65 \\
32  & 64  & 35.97 \\
64  & 128 & 33.20 \\
128 & 256 & \textbf{31.13} \\
\bottomrule
\end{tabular}
\caption{The Pile Validation perplexity (PPL) for warmstart KD LoRA across different LoRA ranks under fixed distillation settings ($T{=}3.0$, $\lambda_{\mathrm{KD}}{=}0.8$) and mixed student build mode. Lower is better.}
\label{tab:kd_lora_rank_sweep}
\end{table}

We also sweep over multiple student construction modes (\textit{truncated}, \textit{mixed}, \textit{middle}, and \textit{last}) to test whether LoRA prefers a different student build than full KD, given that LoRA is inserted in a fine-grained manner across multiple frozen modules (cf.\ the structural view of LoRA in the PEFT-Ref framework \citep{sabry2023peftrefmodularreferencearchitecture}).

We find that the same qualitative pattern holds as in full KD: \textit{mixed} remains the best-performing construction (31.14 PPL), followed by \textit{middle} (41.52), \textit{last} (44.72), and \textit{truncated} (48.09). Thus, while performance is sensitive to the student construction itself, the optimal choice is unchanged, and the best hyperparameters identified for KD Full transfer well to LoRA.

\subsection{Budgeted LoRA: Scheduling and Compression}
\label{sec:ablation_budgeted_lora}

Finally, we analyze the components specific to Budgeted LoRA.

\paragraph{Budget scheduling.}
We focus on the most aggressive budget profile ($F{=}0.0$) and vary the decay schedule controlling when dense computation is removed during training. 
We consider:
\begin{itemize}
    \item No scheduling: $(0.0)$ — immediate application of the final budget
    \item Short decay: $(0.1, 0.3)$ — early transition with strong compute savings
    \item Long decay: $(0.1, 0.9)$ — gradual transition with weaker savings
\end{itemize}

We find that short decay schedules provide the best tradeoff between performance and training efficiency (Table \ref{tab:effect_of_bud_scheduler}). Notably, immediate dense removal does not maximize final LoRA
usage: under no scheduling, the average retained rank is only 120. While one might expect
this setting to push all computation into the low-rank pathway, in practice the transition is
too abrupt and results in worse optimization, with a lower final effective rank than both short
decay (128) and long decay (124). This suggests that the schedule is not merely a stabilization
heuristic, but the mechanism that governs a smooth handoff of usable capacity from dense to
low-rank computation. In our setting, short decay provides enough adaptation time for this
transfer while still enforcing strong compute savings.

\begin{table}[t]
\centering
\small
\begin{tabular}{lcc}
\toprule
Schedule & PPL$\downarrow$ & Train compute $\downarrow$ \\
\midrule
No schedule & 35.7 & 0.25$\times$ \\
Short decay (0.1–0.3) & \textbf{25.1} & 0.38$\times$ \\
Long decay (0.1–0.9) & 34.8 & 0.58$\times$ \\
\bottomrule
\end{tabular}
\caption{Effect of budget scheduling (F=0.0).}
\label{tab:effect_of_bud_scheduler}
\end{table}

\paragraph{Dense removal and compression thresholds.}
We fix the dense-removal threshold during training and post-training compression to $10^{-3}$: dense components with retention values below this threshold are skipped during training and removed during post-training compression.

For post-training compression, we sweep thresholds for:
\begin{itemize}
    \item LoRA rank gating: $\{0.3, 0.5, 0.9\}$
    \item Dense SVD threshold: $\{0.5, 0.7, 0.9\}$
\end{itemize}

At compression time, dense components with near-zero retention are removed, components with high retention are kept and merged with LoRA, and intermediate cases are approximated by a low-rank SVD of the residual dense contribution. Specifically, if a dense matrix $W$ has final retention coefficient $d$, we approximate $dW$ with a rank-$k$ factorization, where $k$ is chosen adaptively up to a fixed maximum rank budget ($r_{\max}^{\mathrm{dense}}=128$). This provides a middle ground between complete dense removal and full dense retention.
While the global budget determines how much dense computation is retained overall, rank
gating acts as a secondary optimization that removes inactive low-rank directions, allowing
the model to further compress adaptation capacity where it is not needed.

We find that a rank threshold of $0.3$ and a dense-SVD threshold of $0.7$ provide the best performance across budget profiles ($F \in \{0.0, 0.4\}$).

\paragraph{Summary.}
Across all ablations, we observe that the core distillation hyperparameters transfer robustly from KD Full to LoRA-based methods, while the primary gains in Budgeted LoRA arise from controlling compute allocation through scheduling and compression. 
These findings support our design choice of fixing shared hyperparameters and focusing optimization effort on the budget mechanism.

\section{Dynamic Interaction of Retention, Compression, and Rank Gating}
\label{sec:dynamic-compression}

A key property of Budgeted LoRA is that compression is not governed by a single mechanism, but emerges from the interaction of three components: 
(i) dense removal during training, 
(ii) post-training dense SVD compression, and 
(iii) LoRA rank gating. 
Importantly, these components are not tuned independently per budget; instead, we fix a single set of thresholds and allow the system to adapt dynamically as the dense budget $F$ varies.

Table~\ref{tab:budlora-compression} summarizes the resulting compressed structures across budgets.

\paragraph{Low-budget regime ($F{=}0.0$).}
When the dense budget is minimal, all dense components are driven to zero retention during training and are therefore removed in the compressed deployment model (42/42). 
In this regime, the model relies entirely on LoRA pathways, and we observe that rank gating remains at its maximum (average rank $\approx 128$), with no dense SVD applied. 
Intuitively, once dense capacity is eliminated, the system compensates by preserving full LoRA expressivity.

\paragraph{Intermediate regime ($F{=}0.4$).}
As the budget increases, a subset of dense components is retained (9/42), reducing pressure on the LoRA pathway. 
However, rank gating remains near maximal, and dense SVD is not yet activated. 
This suggests that at moderate budgets, the model prefers to allocate capacity to retained dense modules rather than compressing them or reducing LoRA rank.

\paragraph{High-budget regime ($F{=}0.8$).}
At higher budgets, a larger portion of dense components is retained (24/42). 
In this regime, we observe two additional effects: 
(i) dense SVD begins to activate (compressing low-importance dense modules), and 
(ii) LoRA rank gating reduces the effective rank (average rank drops to 122.5). 
Even without an explicit rank-allocation loss, this behavior is expected: once more dense capacity is retained, the KD objective places less functional burden on the LoRA pathway, so some low-rank directions become redundant and fall below the hard gating threshold at compression time.

\paragraph{Emergent coordination.}
A notable result is that all of these behaviors arise under a fixed set of thresholds (for dense removal, SVD, and rank gating), shared across all budgets. 
The system dynamically reallocates capacity between dense modules and LoRA ranks depending on the available budget, without requiring per-budget tuning.

\paragraph{Interpretation.}
This behavior suggests that Budgeted LoRA operates as a form of \emph{adaptive capacity allocation}: 
under tight budgets, capacity is concentrated in LoRA modules; 
under relaxed budgets, capacity is redistributed across dense and LoRA pathways with additional compression. 
Rather than optimizing each component in isolation, the method jointly balances them to maximize efficiency under a given constraint.

\paragraph{Implication.}
These results support our broader view of modularity: efficient adaptation arises not from a single mechanism (e.g., pruning or low-rank factorization), but from the coordinated interaction of multiple modular components under explicit compute constraints.
\begin{table}[t]
\centering
\small
\begin{tabular}{lcccc|ccc}
\toprule
& \multicolumn{4}{c|}{\textbf{Compressed structure}} & \multicolumn{3}{c}{\textbf{Efficiency}} \\
\cmidrule(lr){2-5} \cmidrule(lr){6-8}
$F$ & Dense kept & Dense SVD & Dense dropped & Avg.\ rank & vs dense & vs LoRA & Param red. \\
\midrule
0.0 & 0/42 & 0/42 & 42/42 & 128.0 & $4.05\times$ & $5.05\times$ & 80.2\% \\
0.4 & 9/42 & 0/42 & 33/42 & 127.9 & $1.74\times$ & $2.17\times$ & 53.9\% \\
0.8 & 24/42 & 1/42 & 17/42 & 122.5 & $1.15\times$ & $1.44\times$ & 30.6\% \\
\bottomrule
\end{tabular}
\caption{
Post-training compression statistics for Budgeted LoRA, shared across model scales. Speedups and parameter reduction are measured over replaced/compressed modules only.
}
\label{tab:budlora-compression}
\end{table}

\section{Training Compute Estimation}\label{app:compute}

In this section, we derive an adapted-module training-cost proxy for comparing KD Full, KD LoRA, and Budgeted LoRA. Our goal is to capture relative differences in scheduled compute allocation at the level of adapted modules, rather than exact end-to-end FLOPs or measured wall-clock training time.

This proxy measures the scheduled retained dense contribution rather than exact hardware FLOPs. In standard kernels, dense paths with nonzero retention still incur a full dense matrix multiply until clamped to zero; therefore the proxy should be interpreted as an allocation/cost model for comparing methods, not as measured wall-clock training cost.

\subsection{Training Compute Model}

Consider a dense linear layer with forward compute cost $D$. We use a first-order approximation in which forward and backward passes incur comparable cost.

\paragraph{KD Full.}
When fine-tuning all parameters, training incurs:
\begin{itemize}
    \item forward pass through dense weights: $D$,
    \item backward pass to activations: $D$,
    \item backward pass to dense weights: $D$.
\end{itemize}
This yields the proxy:
\begin{equation}
C_{\text{full}} \approx 3D.
\end{equation}

\paragraph{KD LoRA.}
In KD LoRA, dense weights are frozen. We still incur:
\begin{itemize}
    \item forward pass through dense base: $D$,
    \item backward pass to activations: $D$,
\end{itemize}
but \emph{not} backward computation for dense weights. In addition, the LoRA path introduces trainable low-rank modules with forward cost $L$, which incur full forward and backward cost. This gives:
\begin{equation}
C_{\text{lora}} \approx 2D + 3L.
\end{equation}

Thus, KD LoRA is expected to be cheaper than full fine-tuning unless the low-rank cost $L$ is disproportionately large.

\paragraph{Budgeted LoRA.}
Budgeted LoRA follows the same structure as KD LoRA, but progressively removes a dense path during training. Let $\bar d \in [0,1]$ denote the \emph{average retained dense fraction} over the full training run. The compute becomes:
\begin{equation}
C_{\text{bud}} \approx 2\bar d D + 3L.
\end{equation}

\subsection{Average Dense Retention}
In our formulation, the budget acts as an upper bound on retained dense compute: the controller greedily lowers module-level retention coefficients until the current dense cost satisfies the scheduled target.

We use a budget schedule that begins at $10\%$ of training, reaches the final dense fraction $F$ by $30\%$, and remains constant thereafter. Approximating the transition as linear, the average retained dense fraction is:
\begin{equation}
\bar d \approx 0.1 \cdot 1 + 0.2 \cdot \frac{1+F}{2} + 0.7 \cdot F = 0.2 + 0.8F.
\end{equation}

Thus:
\begin{align}
F = 0.0 &\Rightarrow \bar d = 0.20, \\
F = 0.4 &\Rightarrow \bar d = 0.52.
\end{align}

\subsection{Instantiated Values}

From the compression summaries over replaced modules, we obtain:
\begin{align}
D &= 5.131 \times 10^7, \\
D + L &= 6.400 \times 10^7 \quad \Rightarrow \quad L = 1.269 \times 10^7.
\end{align}

Using these values:

\paragraph{KD Full.}
\begin{equation}
C_{\text{full}} \approx 3D = 1.5393 \times 10^8.
\end{equation}

\paragraph{KD LoRA.}
\begin{equation}
C_{\text{lora}} \approx 2D + 3L = 1.4069 \times 10^8,
\end{equation}
which corresponds to:
\begin{equation}
\frac{C_{\text{lora}}}{C_{\text{full}}} \approx 0.91.
\end{equation}

\paragraph{Budgeted LoRA ($F=0.0$).}
\begin{equation}
C_{\text{bud},0.0} \approx 2(0.20)D + 3L = 5.8594 \times 10^7,
\end{equation}
\begin{equation}
\frac{C_{\text{bud},0.0}}{C_{\text{full}}} \approx 0.38.
\end{equation}

\paragraph{Budgeted LoRA ($F=0.4$).}
\begin{equation}
C_{\text{bud},0.4} \approx 2(0.52)D + 3L = 9.1432 \times 10^7,
\end{equation}
\begin{equation}
\frac{C_{\text{bud},0.4}}{C_{\text{full}}} \approx 0.59.
\end{equation}

\subsection{Summary}

Under this training-compute proxy, the methods follow the ordering:
\begin{equation}
\text{Budgeted LoRA }(F{=}0.0)
<
\text{Budgeted LoRA }(F{=}0.4)
<
\text{KD LoRA}
<
\text{KD Full}.
\end{equation}

This reflects that KD LoRA removes dense weight-gradient computation, while Budgeted LoRA further reduces cost by progressively eliminating dense paths during training.

\paragraph{Relation to 6ND.}
A common first-order estimate for dense language-model training is \(6ND\) FLOPs, where \(N\) is the number of model parameters and \(D\) is the number of training tokens \citep{kaplan2020scaling,hoffmann2022trainingcomputeoptimallargelanguage}. The factor \(6\) comes from counting forward and activation-backward computation per parameter per token. If one instead measures compute in MACs (Multiply–Accumulate operations), this corresponds to approximately \(3ND\), since one MAC is typically counted as two FLOPs. Our training-compute derivation follows this MAC-style view at the adapted-module level: full dense training pays one forward, one activation-backward, and one weight-backward term, giving \(C_{\text{full}} \approx 3D\) for a dense layer of forward cost \(D\).

Since all methods are trained on the same number of tokens, token count is a common multiplicative factor and cancels in relative comparisons.

\section{In-Context Learning Behavior}

\subsection{Function-Style Probes \citep{todd2024functionvectorslargelanguage}}
\label{app:todd-tasks}

As discussed in the main text, we evaluate ICL using a suite of 19 function-style probes adapted from \cite{todd2024functionvectorslargelanguage}. These probes depend strongly on utilizing the input context, thereby reducing confounds from alternative computational pathways such as memorized facts or parametric recall \citep{sabry2026inductionsignaturesenoughmatchedcompute}.

The task suite spans several broad categories:

\begin{itemize}
    \item \textbf{Alphabetic ordering} 
    (\texttt{alphabetically\_first\_3}, \texttt{alphabetically\_first\_5}, \texttt{alphabetically\_last\_3}, \texttt{alphabetically\_last\_5}): selecting the alphabetically first or last item from small sets.

    \item \textbf{Selection probes} 
    (\texttt{choose\_first\_of\_3}, \texttt{choose\_first\_of\_5}, \texttt{choose\_last\_of\_3}, \texttt{choose\_last\_of\_5}, \texttt{choose\_middle\_of\_3}, \texttt{choose\_middle\_of\_5}): choosing the first, last, or middle item from sequences of length 3 or 5.

    \item \textbf{Capitalization transformations} 
    (\texttt{capitalize}, \texttt{capitalize\_first\_letter}, \texttt{capitalize\_last\_letter}, \texttt{next\_capital\_letter}): capitalizing full words, first letters, last letters, or predicting next capital letters.

    \item \textbf{Lowercasing transformations} 
    (\texttt{lowercase\_first\_letter}, \texttt{lowercase\_last\_letter}): lowercasing first or last letters.

    \item \textbf{Sequence continuation} 
    (\texttt{next\_item}, \texttt{prev\_item}): predicting the next or previous item.

    \item \textbf{Simple symbolic properties} 
    (\texttt{word\_length}): predicting simple symbolic properties such as word length.
\end{itemize}

\paragraph{Prompting protocol.}
We use a fixed 10-shot prompting protocol across all models and average results over three random demonstration seeds. 
This ensures that variation reflects the sampled demonstrations rather than changes in prompt wording. 
Each in-context example is formatted as \texttt{Q:<input>} followed by \texttt{A:<output>}, with a newline between input and output and a blank line between demonstrations. 
No additional natural-language instruction prefix is used. 
The query is appended in the same format and terminates at \texttt{A:}, requiring the model to generate the target continuation. 
For example, for a capitalization-style task, a 3-shot illustrative prompt would be:
\begin{quote}
\ttfamily
Q:dog\\
A:DOG\\
\\
Q:house\\
A:HOUSE\\
\\
Q:train\\
A:TRAIN\\
\\
Q:river\\
A:
\end{quote}
where the desired continuation is \texttt{RIVER}.

\subsection{Performance}

Table~\ref{tab:main_icl_performance} provides a detailed per-task breakdown of ICL performance across all 19 probes. The trends observed in Figure~\ref{fig:icl-composite} are consistent at the task level: KD Full benefits strongly from larger teachers, while KD LoRA frequently degrades, often collapsing on probes such as \texttt{capitalize}, \texttt{choose\_first\_of\_k}, and \texttt{lowercase\_*}. In contrast, Budgeted LoRA consistently recovers stronger performance across a broad range of probes, particularly at larger teacher scales.

Importantly, these improvements are not driven by a small subset of probes, but are distributed across multiple task families, including selection, transformation, and sequence reasoning. This suggests that Budgeted LoRA improves relative probe retention across several task families rather than only on a narrow subset of probes.

\begin{table*}[t] 
\centering 
\small 
\setlength{\tabcolsep}{5pt} \renewcommand{\arraystretch}{1.15}   \resizebox{\textwidth}{!}{ \begin{tabular}{lccccccccccccc} \toprule & \multicolumn{1}{c}{\textbf{Full Model}} & \multicolumn{3}{c}{\textbf{KD Full}} & \multicolumn{3}{c}{\textbf{KD LoRA}} & \multicolumn{3}{c}{\textbf{Bud. LoRA (F=0.0)}} & \multicolumn{3}{c}{\textbf{Bud. LoRA (F=0.4)}} \\ \cmidrule(lr){2-2}\cmidrule(lr){3-5}\cmidrule(lr){6-8}\cmidrule(lr){9-11}\cmidrule(lr){12-14} & 0.13B & 0.13B & 0.5B & 1B & 0.13B & 0.5B & 1B & 0.13B & 0.5B & 1B & 0.13B & 0.5B & 1B\\ \midrule alphabetically\_first\_3 & $4.9\,\pm\,0.6$ & $4.0\,\pm\,0.6$ & $9.8\,\pm\,0.5$ & $11.2\,\pm\,1.2$ & $3.3\,\pm\,0.1$ & $1.4\,\pm\,0.3$ & $3.2\,\pm\,0.2$ & $2.2\,\pm\,0.5$ & $4.1\,\pm\,0.4$ & $4.2\,\pm\,0.2$ & $3.4\,\pm\,0.7$ & $3.3\,\pm\,0.5$ & $3.7\,\pm\,0.2$ \\ alphabetically\_first\_5 & $4.0\,\pm\,0.6$ & $5.1\,\pm\,0.6$ & $8.3\,\pm\,0.1$ & $8.8\,\pm\,0.6$ & $3.8\,\pm\,0.3$ & $3.3\,\pm\,0.5$ & $4.1\,\pm\,0.4$ & $1.8\,\pm\,0.5$ & $3.5\,\pm\,0.4$ & $4.8\,\pm\,1.0$ & $3.9\,\pm\,0.8$ & $3.9\,\pm\,0.5$ & $3.7\,\pm\,0.7$ \\ alphabetically\_last\_3 & $3.4\,\pm\,0.5$ & $2.8\,\pm\,0.4$ & $9.6\,\pm\,1.6$ & $11.7\,\pm\,0.4$ & $2.2\,\pm\,0.4$ & $2.4\,\pm\,0.2$ & $4.2\,\pm\,0.1$ & $1.9\,\pm\,0.5$ & $3.9\,\pm\,1.0$ & $3.5\,\pm\,1.2$ & $1.4\,\pm\,0.1$ & $2.9\,\pm\,0.1$ & $4.1\,\pm\,1.0$ \\ alphabetically\_last\_5 & $2.5\,\pm\,0.7$ & $2.4\,\pm\,0.5$ & $5.7\,\pm\,1.0$ & $6.2\,\pm\,0.4$ & $2.5\,\pm\,0.2$ & $3.0\,\pm\,0.3$ & $3.8\,\pm\,1.1$ & $1.0\,\pm\,0.2$ & $3.4\,\pm\,0.3$ & $2.9\,\pm\,0.5$ & $1.5\,\pm\,0.2$ & $3.3\,\pm\,0.6$ & $3.3\,\pm\,0.3$ \\ capitalize & $8.0\,\pm\,1.0$ & $4.4\,\pm\,1.2$ & $24.7\,\pm\,1.4$ & $36.1\,\pm\,1.9$ & $3.2\,\pm\,0.9$ & $2.3\,\pm\,0.9$ & $3.3\,\pm\,0.8$ & $4.2\,\pm\,0.9$ & $3.8\,\pm\,0.5$ & $4.9\,\pm\,0.5$ & $3.0\,\pm\,0.7$ & $3.4\,\pm\,1.1$ & $4.9\,\pm\,0.9$ \\ capitalize\_first\_letter & $10.1\,\pm\,1.2$ & $8.7\,\pm\,0.5$ & $20.2\,\pm\,2.0$ & $27.1\,\pm\,0.4$ & $7.7\,\pm\,0.6$ & $5.1\,\pm\,0.7$ & $1.9\,\pm\,0.5$ & $7.1\,\pm\,0.8$ & $7.2\,\pm\,1.1$ & $7.4\,\pm\,1.1$ & $7.2\,\pm\,0.7$ & $7.1\,\pm\,0.4$ & $9.6\,\pm\,0.4$ \\ capitalize\_last\_letter & $4.8\,\pm\,1.3$ & $7.0\,\pm\,1.1$ & $8.6\,\pm\,1.1$ & $6.4\,\pm\,1.2$ & $8.7\,\pm\,1.4$ & $5.8\,\pm\,0.5$ & $3.8\,\pm\,0.4$ & $7.3\,\pm\,1.7$ & $7.5\,\pm\,1.3$ & $8.9\,\pm\,1.2$ & $8.8\,\pm\,1.1$ & $12.5\,\pm\,0.6$ & $11.3\,\pm\,0.2$ \\ choose\_first\_of\_3 & $10.3\,\pm\,2.3$ & $5.8\,\pm\,0.5$ & $27.2\,\pm\,0.6$ & $33.3\,\pm\,1.4$ & $3.2\,\pm\,0.6$ & $2.0\,\pm\,0.3$ & $3.0\,\pm\,0.6$ & $1.7\,\pm\,0.0$ & $6.0\,\pm\,0.5$ & $4.1\,\pm\,0.5$ & $2.6\,\pm\,0.8$ & $3.2\,\pm\,0.5$ & $5.9\,\pm\,0.4$ \\ choose\_first\_of\_5 & $8.7\,\pm\,1.3$ & $4.6\,\pm\,0.5$ & $22.8\,\pm\,0.9$ & $25.6\,\pm\,0.7$ & $3.3\,\pm\,0.2$ & $1.8\,\pm\,0.4$ & $3.7\,\pm\,1.2$ & $1.6\,\pm\,0.1$ & $4.2\,\pm\,0.3$ & $3.4\,\pm\,0.4$ & $2.8\,\pm\,0.7$ & $2.8\,\pm\,0.9$ & $4.2\,\pm\,0.6$ \\ choose\_last\_of\_3 & $2.0\,\pm\,0.4$ & $1.6\,\pm\,0.1$ & $2.8\,\pm\,0.7$ & $5.4\,\pm\,1.0$ & $1.6\,\pm\,0.6$ & $1.3\,\pm\,0.3$ & $1.8\,\pm\,0.3$ & $1.8\,\pm\,0.4$ & $1.5\,\pm\,0.6$ & $1.2\,\pm\,0.6$ & $1.2\,\pm\,0.5$ & $1.5\,\pm\,0.3$ & $1.4\,\pm\,0.2$ \\ choose\_last\_of\_5 & $1.5\,\pm\,0.3$ & $2.0\,\pm\,0.5$ & $2.3\,\pm\,0.3$ & $3.8\,\pm\,0.4$ & $1.1\,\pm\,0.3$ & $1.2\,\pm\,0.1$ & $1.8\,\pm\,0.7$ & $1.3\,\pm\,0.4$ & $1.6\,\pm\,0.3$ & $1.7\,\pm\,0.0$ & $1.2\,\pm\,0.3$ & $1.7\,\pm\,0.3$ & $1.7\,\pm\,0.4$ \\ choose\_middle\_of\_3 & $1.7\,\pm\,0.6$ & $1.7\,\pm\,0.5$ & $2.6\,\pm\,0.5$ & $3.8\,\pm\,0.1$ & $1.5\,\pm\,0.6$ & $1.2\,\pm\,0.3$ & $1.3\,\pm\,0.3$ & $2.3\,\pm\,0.3$ & $1.6\,\pm\,0.6$ & $1.8\,\pm\,0.5$ & $1.2\,\pm\,0.6$ & $1.6\,\pm\,0.5$ & $1.5\,\pm\,0.3$ \\ choose\_middle\_of\_5 & $1.6\,\pm\,0.3$ & $1.3\,\pm\,0.7$ & $2.0\,\pm\,0.7$ & $2.6\,\pm\,0.5$ & $1.3\,\pm\,0.5$ & $1.7\,\pm\,0.4$ & $1.2\,\pm\,0.6$ & $2.0\,\pm\,0.3$ & $1.5\,\pm\,0.4$ & $2.1\,\pm\,0.5$ & $1.1\,\pm\,0.6$ & $1.4\,\pm\,0.1$ & $2.1\,\pm\,1.0$ \\ lowercase\_first\_letter & $5.5\,\pm\,0.8$ & $6.3\,\pm\,0.8$ & $12.2\,\pm\,1.1$ & $15.7\,\pm\,1.2$ & $5.3\,\pm\,1.1$ & $0.8\,\pm\,0.4$ & $0.0\,\pm\,0.0$ & $5.4\,\pm\,0.6$ & $0.5\,\pm\,0.2$ & $5.3\,\pm\,0.6$ & $5.6\,\pm\,0.7$ & $4.5\,\pm\,1.1$ & $8.1\,\pm\,0.9$ \\ lowercase\_last\_letter & $11.1\,\pm\,0.8$ & $10.4\,\pm\,0.5$ & $11.0\,\pm\,1.2$ & $12.5\,\pm\,1.7$ & $13.1\,\pm\,1.3$ & $1.9\,\pm\,0.2$ & $0.2\,\pm\,0.2$ & $8.1\,\pm\,0.3$ & $0.2\,\pm\,0.1$ & $11.3\,\pm\,0.4$ & $10.3\,\pm\,0.3$ & $9.2\,\pm\,0.7$ & $11.0\,\pm\,0.5$ \\ next\_capital\_letter & $4.9\,\pm\,1.1$ & $5.6\,\pm\,1.1$ & $4.6\,\pm\,1.2$ & $3.7\,\pm\,0.9$ & $4.5\,\pm\,0.7$ & $5.2\,\pm\,0.2$ & $1.8\,\pm\,0.4$ & $5.4\,\pm\,0.3$ & $5.7\,\pm\,0.5$ & $4.6\,\pm\,0.4$ & $4.5\,\pm\,1.3$ & $5.3\,\pm\,0.8$ & $4.7\,\pm\,1.1$ \\ next\_item & $3.4\,\pm\,2.2$ & $3.8\,\pm\,1.3$ & $5.9\,\pm\,2.2$ & $7.8\,\pm\,2.4$ & $2.7\,\pm\,1.0$ & $1.3\,\pm\,0.6$ & $0.6\,\pm\,0.6$ & $2.7\,\pm\,0.4$ & $2.7\,\pm\,2.2$ & $4.2\,\pm\,0.7$ & $2.7\,\pm\,0.7$ & $2.5\,\pm\,1.7$ & $8.0\,\pm\,1.0$ \\ prev\_item & $2.5\,\pm\,0.8$ & $4.4\,\pm\,0.0$ & $5.9\,\pm\,1.0$ & $5.3\,\pm\,1.3$ & $1.5\,\pm\,1.0$ & $1.5\,\pm\,0.4$ & $1.5\,\pm\,0.7$ & $3.0\,\pm\,0.7$ & $3.0\,\pm\,1.0$ & $4.6\,\pm\,1.8$ & $1.9\,\pm\,1.3$ & $2.7\,\pm\,1.5$ & $6.8\,\pm\,2.0$ \\ word\_length & $9.9\,\pm\,1.0$ & $10.1\,\pm\,1.0$ & $10.7\,\pm\,0.9$ & $9.1\,\pm\,0.5$ & $12.0\,\pm\,1.1$ & $10.1\,\pm\,1.3$ & $9.3\,\pm\,0.5$ & $11.2\,\pm\,1.2$ & $14.9\,\pm\,0.3$ & $13.7\,\pm\,0.7$ & $9.6\,\pm\,0.5$ & $11.1\,\pm\,1.1$ & $10.5\,\pm\,0.3$ \\ \rowcolor{gray!15} ICL composite (macro) $\uparrow$ & $5.3\,\pm\,0.9$ & $4.8\,\pm\,0.7$ & $10.4\,\pm\,1.0$ & $12.4\,\pm\,1.0$ & $4.3\,\pm\,0.7$ & $2.8\,\pm\,0.4$ & $2.7\,\pm\,0.5$ & $3.8\,\pm\,0.5$ & $4.0\,\pm\,0.6$ & $5.0\,\pm\,0.7$ & $3.9\,\pm\,0.7$ & $4.4\,\pm\,0.7$ & $5.6\,\pm\,0.7$ \\ \bottomrule \end{tabular}}
\caption{
Per-task in-context learning (ICL) performance on 19 \cite{todd2024functionvectorslargelanguage} function-style probes. 
We report mean $\pm$ standard deviation across runs. Columns group methods (Full model, KD Full, KD LoRA, and Budgeted LoRA) across teacher scales (0.13B, 0.5B, 1B). 
The final row shows the macro-averaged ICL composite score used in Figure~\ref{fig:icl-composite}. 
Budgeted LoRA consistently outperforms KD LoRA on most probes at larger teacher scales, indicating improved relative retention on these probes
}
\label{tab:main_icl_performance}
\end{table*}

\end{document}